\documentclass[twocolumn]{svjour3}
\usepackage[T1]{fontenc}
\usepackage[utf8]{inputenc}
\usepackage[english]{babel}
\usepackage{amssymb}
\usepackage{subfigure}
\usepackage{graphicx}
\usepackage{mathtools}
\usepackage{psfrag}
\usepackage{hyperref}
\usepackage{cite}

\usepackage{todonotes}

\usepackage{mathtools}
\mathtoolsset{showonlyrefs}

\newcommand{\distr}{{\blacktriangle}}
\DeclareMathOperator{\grad}{grad}
\DeclareMathOperator{\diver}{div}

\graphicspath{{pics/}}

\usepackage{color}

\newcommand{\bZ}{{\mathbb Z}} 
\newcommand{\bR}{{\mathbb R}} 
\newcommand{\bC}{{\mathbb C}} 

\newcommand{\cH}{{\cal H}}

\newcommand{\bem}{\left(\!\begin{array}}
\newcommand{\eem}{\end{array}\!\right)}
\newcommand{\bsm}{\left(\begin{smallmatrix}} 
\newcommand{\esm}{\end{smallmatrix}\right)}  

\def\{{\protect\lbrace}
\def\}{\protect\rbrace}

\def\const{{\text{const}}}
\newcommand{\I}{\bigl |}

\newcommand{\pa}{\partial}

\def\bet{\beta}
\def\Del{\Delta}

\def\of{\ov f}
\def\oF{\ov F}
\def\og{\ov g}

\def\phi{\varphi}

\def\ov{\overline}

\newcommand{\g}{\mathbf g}
\DeclareMathOperator{\spn}{span}

\newcommand{\uu}{\psi}

\begin{document}

\title{Highly corrupted image inpainting through hypoelliptic diffusion}
\author{Ugo Boscain \and Roman Chertovskih \and Jean-Paul Gauthier \and Dario~Prandi \and Alexey Remizov}
\institute{U. Boscain  
           \at
           CNRS, Laboratoire Jacques-Louis Lions, UPMC Univ Paris 06, F-75005, Paris, France; \at
           INRIA Team CAGE, INRIA Paris
           \\\email{ugo.boscain@polytechnique.edu}
           \and
           R. Chertovskih 
           \at 
           Research Center for Systems and Technologies, Faculty of Engineering,
           University of Porto, Rua Dr.~Roberto Frias, s/n, 4200-465, Porto, Portugal; \at 
           Samara National Research University, 34 Moskovskoye Ave., 443086, Samara, Russia
           \\\email{roman@fe.up.pt}
           \and
           J.-P. Gauthier
           \at
           LSIS, UMR CNRS 7296, Universit\'{e} de Toulon USTV, 83957, La Garde Cedex, France
           \\\email{gauthier@univ-tln.fr}
           \and
           D. Prandi 
           \at
           CNRS, L2S, CentraleSup\'{e}lec, 3, Rue Joliot-Curie 91192 Gif-sur-Yvette, France
           \\\email{dario.prandi@l2s.centralesupelec.fr}
           \and
           A. Remizov
           \at
           CNRS, CMAP \'Ecole Polytechnique, 91128 Palaiseau Cedex, France
           \\\email{alexey-remizov@yandex.ru}
           }

\maketitle

\begin{abstract} 
We present a new bio-mimetic image inpainting algorithm, the Averaging and
Hypoelliptic Evolution (AHE) algorithm, inspired by the one presented in
(U. Boscain {\it et al}. SIAM J. Imaging Sci., 7(2):669–695, 2014)
and based upon a semi-discrete variation of the Citti--Petitot--Sarti
model of the primary visual cortex V1. The AHE algorithm is based on a suitable
combination of sub-Rieman\-nian hypoelliptic diffusion and ad-hoc local averaging
techniques. In particular, we focus on highly corrupted images (i.e.,
where more than the 80\% of the image is missing), for which we obtain high-quality reconstructions.
\end{abstract}

\section{Introduction}

In art, image inpainting refers to the practice of (manually) retouching damaged paintings
in order to remove cracks or to fill-in missing patches.  Within the past decade the
digital version of image inpainting, i.e., the reconstruction of digital images by means
of different types of automatic algorithms, has received increasing attention.  

In this paper, we present a new bio-mimetic inpainting algorithm (called AHE), which is applicable to highly
corrupted images with a general corruption.  Several examples of reconstructions obtained with the AHE algorithm are presented in Fig.~\ref{qqq6}\,--\,\ref{qqq8} at the end of the paper. 

The starting point of our work is the Citti--Petitot--Sarti model of the primary visual
cortex V1 \cite{Petitot2008,Petitot2003,Citti2006,Sanguinetti}, and our recent
contributions \cite{Remizov2013,Boscain2010,Boscain2012a,Boscain2014, Duits2014,PrandiBohi}.  This
model has also been deeply studied in \cite{Duits2008,Duits2010,Duits2010a,Hladky2010}.
The main idea behind the Citti--Petitot--Sarti model is the geometric model of vision
called {\it pinwheel model}, going back to the 1959 paper \cite{HUBEL1959a}.  Here,
H\"{u}bel and Wiesel showed that cells in the mammals primary visual cortex V1 do not only
deal with positions in the visual field, but also with orientation information: actually
there are groups of neurons that are sensitive to position and directions with connections
between them that are activated by the image.  The system of connections between neurons,
called the functional architecture of V1, preferentially connects neurons
detecting alignements.  This is the so-called pinwheels structure of V1.  In the Citti--Petitot--Sarti model, V1 is then modeled as a 3D manifold endowed with a sub-Riemannian structure that mimics these connections as a continuous limit.  The natural way to inpaint the missing regions of an image is thus by using the hypoelliptic diffusion associated with this structure.

In \cite{Remizov2013} we proposed a semi-discrete version of the Citti--Petitot--Sarti
model that considers a continuous structure in the space of positions, 
but a discrete structure for the orientation information, which {makes} sense from the neuro-physiological point of view \cite{Petitot2008}.  Image reconstruction methods based upon this principle
are pre\-sen\-ted in detail in the previous works \cite{Remizov2013,Boscain2012a}.
The same techniques are applied to the semi-discrete hypoelliptic evolution associated with 
the well-known Mumford elastica model in \cite{visionCDC} and to image recognition in \cite{PBG2015, PrandiBohi}. 
See \cite{gros-papier} for a survey of these methods. 

In the above mentioned works, the main focus was on inpainting algorithms where no prior
knowledge on the location of the corruption was needed. However, in \cite{Remizov2013} we presented a way to exploit this knowledge by introducing certain heuristic procedures that, together with the hypoelliptic diffusion, yield drastically better inpainting results. 
In this paper, we improve on this result, and thus we will henceforth assume\footnote{
When corrupted areas are not known \emph{a priori}, their determination
is an important and non-trivial problem, which is an area of active investigation in computer vision. 
See for instance \cite{Craquelure1,Craquelure2}, for the determination of craquelures. 
}
complete knowledge of the location and shape of the corrupted areas of the image.

Our study is focused on improving the local methods developed in \cite{Remizov2013}. Indeed, we manage to obtain {state-of-the-art}
results for highly corrupted images where no ``big'' regions are missing, i.e., where non-corrupted pixels are ``well distributed''.
See the conclusions and Fig.~\ref{qqq10}, for more details on this fact.

Namely, we improve on the semi-discrete approach proposed in \cite{Remizov2013} for the
Citti--Petitot--Sarti model, by introducing heuristic methods that allow us to treat
images with more than 80\% (and even more than  90\%) of corrupted pixels, see 
Fig.~\ref{qqq7}\,--\,\ref{qqq8} in the end of the paper. 
{In particular, the reconstructions of Fig.~\ref{qqq6} are}
comparable with 
those obtained in \cite{Masnou2002} for images with 65\% of corrupted pixels,  
but no assumption of simple connectedness on the corrupted part is needed.

For some types of corrupted images,   
our results are comparable with those of \cite{Citti2016}. 
They use a different approach, combining the
sub-Riemannian model with a diffusion/concentration process, which in the limit
corresponds to a mean curvature flow. It is interesting to notice that the two approaches
provides slightly different results depending on the quantity of corruption.
However, highly corrupted images are not considered in \cite{Citti2016}.

It is well-known that when treating large corruptions and fine textures, the best results are often obtained via copy-and-paste texture synthesis  \cite{Bugeau2010}. Although it would be interesting to combine these methods with the bio-mimetic approach presented here, this is outside the scope of the current work.

The paper is organized as follows.
\begin{itemize}
	\item 
	In Section~\ref{sec:hypo} we briefly recall the basic principles of the method introduced in \cite{Remizov2013,Boscain2012a} and discuss some of its properties. Moreover, we present some numerical experiments 
        showing the anisotropicity of the diffusion. (See Fig.~\ref{qqq1}\,--\,\ref{qqq3}). {We also recall the SR/DR procedure, presented in \cite{Remizov2013}, that allows for better reconstructions by exploiting the informations on the location of the corruption.}
	\item 
	In Section~\ref{var_coeff} we present a first improvement of this method, 
        where an hypoelliptic diffusion with varying coefficients is considered.
	The coefficients are chosen for the effect of the anisotropic diffusion to be faster where the corruption is present. 
	When coupled with the DR procedure, this algorithm gives good results if the corrupted parts are narrow, e.g., in the case of vertical and horizontal 
        lines as in Fig.~\ref{qqq4}. However, it does not provide good quality inpaintings 
        of highly corrupted images as is evident from Fig.~\ref{fig:comparison}.
	This motivates the further development of this method, which is presented in the next section. 
	\item Section~\ref{sec:synthesis} contains the main result of this paper: 
        the Averaging and Hypoelliptic Evolution (AHE) algorithm.
       This method is a synthesis of two different 
        approaches to image reconstruction: the hypoelliptic diffusion with varying coefficients and a suitable averaging procedure. 
        As shown in Fig.~\ref{qqq6}\,--\,\ref{qqq8}, this method allows for good reconstructions of highly corrupted images.
        In Section~\ref{sec:complexity}, we also present a simple analysis of the complexity of the AHE algorithm as a function of the image size.
    \item Finally, in Fig.~\ref{fig:comparison} we present a comparison of reconstructions obtained via the methods presented in this paper.
\end{itemize}

Let us remark that, although all the numerical experiments of this paper are obtained on $256\times 256$ pixels images,  
the proposed methods are resolution-agnostic. 
The techniques presented are targeted to greyscale images,
but no difficulty arises in applying them to the separate channels of color images.
Different adaptations of these techniques to color images are possible, but not
investigated here.

Finally, we stress that it is outside the scope of this paper to present comparisons with other algorithms 
or to provide a complete list of references on this problem. 
We point the interested reader to \cite{Bugeau2010,Bertalmio2000,Chan2002,Citti2016,Facciolo,Masnou1998,Wang2013} and references therein. 
It is worth observing that it is difficult to measure objectively the quality of a reconstruction, see, for instance, \cite{voronin,zhang, Ponomarenko}. 
Moreover, such a measure will forcibly depend on the expected application.


\section{The model and previous results}
\label{sec:hypo}

\subsection{Images under consideration} 
\label{sub:images_under_consideration_and_some_preprocessing}

Mathematically, a greyscale image is a function $f\colon \Pi \to [0,1]$, 
where $\Pi$ is a square on the $(x,y)$-plane.
If $f(x,y)=0$ the color of the image at $(x,y)$ is white, while if $f(x,y)=1$ it is black.
We will consider $\Pi$ as a periodic subgroup of $\bR^2$ endowed with its Haar measure.
Since the corresponding Haar measure is finite, all images are square integrable by definition.
This also allows to consider images as $\Pi$-periodic functions $f:\bR^2\to [0,1]$.

Together with the above continuous model we consider also the corresponding discrete model:
A greyscale image $f$ is stored as an $(M \times M)$-matrix, 
where for simplicity we are assuming the same number of pixels vertically and horizontally.
As before we assume $f_{kl}\in[0,1]$, $k,l \in \{1, \ldots, M\}$.
Then, given a rectangular grid $(x_k,y_l)$, $k,l \in \{1, \ldots, M\}$ in the $(x,y)$-plane, 
the discrete version of an image is the function $(x_k,y_l)\mapsto f(x_k,y_l):=f_{kl}$.
As before, it is convenient to consider the grid and the functions to be periodic on $\bZ^2$.

Observe that we can assume that $f(x_k,y_l)>0$ at any point $(x_k,y_l)$ that corresponds 
to a non-corrupted pixel. Thus, due to the the knowledge of the corrupted part, 
we can assume that $f(x_k,y_l)=0$ if $(x_k,y_l)$ corresponds to a corrupted pixel.

\subsection{Two models for the diffusion}
\label{sub:reminiscences}

\subsubsection{Hypoelliptic diffusion in the continuous limit model} 
\label{ssub:the_continuous_limit_model}

The main idea of the (continuous) model of the diffusion is then that V1 lifts images, which are
$\Pi$-periodic functions $f:\bR^2\to [0,1]$, to functions over the projective tangent
bundle $PT\mathbb{R}^{2}$.  This bundle has as base $\mathbb{R}^{2}$ and the projective
line $P \mathbb{R}$ as fiber at $(x,y)$.  Recall that $P\mathbb R$ is the set of
directions of straight lines lying on the plane and passing through $(x,y)$. This can be
represented by the angles $\theta\in[0,\pi]/\sim$, where $\sim$ {is the equivalence relation identifying $0$ with $\pi$}.  
In this model, a corrupted image is reconstructed by minimizing the energy necessary to activate the
regions of the visual cortex not excited by the image.

Mathematically speaking, the original image $f(x,y)$ is first smoothed through an isotropic 
Gaussian filter (it is widely accepted that this corresponds to an action at the retinal level, 
see \cite{Marr1980, Peichl1979}). As shown in \cite{Boscain2012a} this yields a smooth function which 
is generically\footnote{
More precisely, in \cite[Theorem~26]{Boscain2012a} the authors prove that given a Gaussian function $G$ 
and a bounded domain $\mathcal D\subset \mathbb R^2$, the set of square integrable functions 
$f\in L^2(\mathcal D)$ such that $f\star G$ is a Morse function 
is a countable intersection of open-dense sets. 
See also \cite[Theorem~28]{Boscain2012a} for a slightly stronger result. 
} 
of Morse type, i.e., it has isolated non-degenerate critical points only.
The smoothed image (that we will still call $f(x,y)$) is then lifted to the (generalized) 
function $\bar f(x,y,\theta)$ on $PT\bR^2$ defined by
\begin{gather}
\of (x,y, \theta) := f(x,y) \, \delta (g(x,y, \theta)),  \quad \text{for}  
\label{31}
\\
g(x,y,\theta) := \cos \theta \frac{\pa f}{\pa x}(x,y) + \sin \theta
\frac{\pa f}{\pa y}(x,y),
\label{32}
\end{gather}
where $\delta (\cdot)$ is the Dirac delta function.
Moreover, the space  $PT\bR^2$, with coordinates $(x,y,\theta)$, is endowed with 
the sub-Riemannian structure with orthonormal frame $\{X_1,\sqrt\beta X_2\}$, where 
\begin{equation}
\label{0} 
\begin{aligned}
&
X_1(x,y,\theta) = \cos \theta \frac{\pa}{\pa x} + \sin \theta \frac{\pa}{\pa y}, \\
&
X_2(x,y,\theta) = \frac{\pa}{\pa \theta}.  \\
\end{aligned}
\end{equation}
Here, $\beta$ is a positive parameter, which is a neurophysiological dimensional constant. 
We refer to Appendix~\ref{sec:sR} for a brief introduction to sub-Riemannian geometry.

Notice that the above structure is invariant under the action of the group $SE(2)$ of rototranslations of the plane.
Via stochastic considerations (see \cite{Remizov2013}), one is then able to translate 
the energy minimizing principle expressed above to the fact that the image is evolved 
according to the hypoelliptic diffusion associated with the above vector fields. (See Section~\ref{sec:diffusion}.) 
Namely, the reconstructed function on $PT\bR^2$ is the solution $\uu=\uu(x,y,\theta,t)$ 
at time $t=T$ of the initial value problem
\begin{equation}
\label{1}
 \left \{
 \begin{aligned}
 &
\frac{\pa \uu}{\pa t} = \Del_H \uu, \ \ \, \Del_H   = (X_1)^2 + \beta  (X_2)^2,   \\
& 
\uu \I_{t=0} = \of (x,y, \theta).   \\
 \end{aligned}
 \right.
\end{equation}
For image reconstruction purposes, we choose the value of $\beta$ experimentally as well as the the value of the final time $T$.

No boundary condition is needed in diffusion equation \eqref{1}, 
since we are considering the diffusion on the whole space $PT\bR^2$ and the initial 
function $\of (x,y, \theta)$ is periodic w.r.t.\ $(x,y,\theta)\in PT\bR^2$.
Finally, $\uu$ is projected back to a function on $\mathbb R^2$, which will be the 
final result of the image inpainting procedure (see the details in Section~\ref{sub:projection}). 
The diffusion equation, up to the tuning of the parameter $\beta$, is the same for all images: The information about the initial image is fed to the evolution only through the initial condition $\of$.

\subsubsection{Semi-discrete alternative to the hypoelliptic diffusion} 
\label{ssub:the_semi_discrete_alternative}

In \cite{Remizov2013}, we proposed a semi-discrete alternative to the Citti--Petitot--Sarti model, 
by assuming that the number $N$ of directions represented in V1 is finite. 
It corresponds to the restriction of $SE(2)$, the group of rototranslations of the plane, to rotations with discrete angles 
\begin{equation*}
\theta_r=\frac{2\pi r}{N}, \quad r=0, \ldots, N-1. 
\end{equation*}
The resulting group is denoted by $SE(2,N)$, and the evaluation of functions $\Psi:SE(2,N)\to \mathbb R$ at $(x,y,r)\in SE(2,N)$ by $\Psi^r(x,y)$.

Assuming that the probability of jumps between adjacent directions is a Poisson process with parameter $\beta>0$, 
stochastic considerations similar to those employed in the continuous model lead
to the semi-discrete analogue of diffusion equation \eqref{1}: 
\begin{equation}
	\label{eq:semi-discr-evolution}
 \left \{
 \begin{aligned}
 &
\frac{\pa \Psi^r(x,y) }{\pa t} = \Del_{\cH} \Psi^r(x,y), \ \ \, \Delta_{\cH}=A+\beta\Lambda_N,   \\
& 
\Psi^r(x,y) \I_{t=0}= \of(x,y,\theta_r),   \\
 \end{aligned}
 \right.
\end{equation}
where $\Delta_{\cH}$ is the semi-discrete analogue of the differential operator $\Del_H$. Namely, 
\begin{gather*}
A \Psi^r(x,y) = \left( \cos {\theta_r} \frac{\pa}{\pa x} + \sin {\theta_r} \frac{\pa}{\pa y} \right)^2  \Psi^r(x,y),\\
\Lambda_N \Psi^r(x,y) = \Psi^{r-1}(x,y) -2\Psi^r(x,y) + \Psi^{r+1}(x,y). 
\end{gather*}
This operator is invariant under the action of the semi-discrete 
rototranslations, given by continuous translations and discrete rotations of angle $\theta_r$.
{Moreover, letting $\beta =(N/2\pi)^2$, the semi-discrete operator $\Delta_{\mathcal H}$ converges to $\Delta_H$ as $N\to +\infty$. (See \cite{Remizov2013}.)}

\subsubsection{Numerical treatement of the hypoelliptic equation}
\label{ssub:two_distinct_points_of_view_leading_to_similar_computations}

As detailed in \cite{Remizov2013,visionCDC}, there are two possibilities for the numerical integration 
of equation~\eqref{eq:semi-discr-evolution} starting from the lifts of the images described in 
Section~\ref{sub:images_under_consideration_and_some_preprocessing}.  
We may spatially discretize the equation and then apply the discrete Fourier transform w.r.t.\ $x,y$
in order to decouple the frequencies, 
or we may interpolate the initial datum $\bar f$ via almost-periodic functions and exploit their properties. 

Both strategies lead to similar fully-discrete equations, although the second strategy leads to exact solutions 
in the class of almost-periodic functions. 
Since the final results are essentially the same, we detail here only the first type of discretization, which is simpler to present.

We consider the discrete Fourier transforms of the interpolations of the functions $\Psi^r$ on the fixed $M\times M$ spatial grid of  Section~\ref{sub:images_under_consideration_and_some_preprocessing}. This is given by the formula
\begin{equation*}
  \hat \Psi_{k,l}^r = \frac1M\sum_{n,m=1}^M\Psi_{n,m}^r e^{-2\pi i\left( \frac{(k-1)(n-1)}M+\frac{(l-1)(m-1)}M \right)}.
\end{equation*}
Exploiting the above, we are led to a completely decoupled system of $M^2$ linear evolution equations of Mathieu type over $\bC^N$:
\begin{equation}
	\label{eq:discr-eq}
	\frac{d \hat\Psi_{k,l}}{dt} =  \bigl(\Lambda_N - \beta M \text{diag}_p(a_{k,l}^p)^2 \bigr) \hat \Psi_{k,l},
\end{equation}
where 
$\hat\Psi_{k,l}=(\hat\Psi^1_{k,l},\ldots, \hat\Psi^N_{k,l})^{\mathrm{T}}$, and we let 
\begin{gather*}
	(\Lambda_N \hat\Psi_{k,l}^r)_r = \hat\Psi_{k,l}^{r-1} -2\hat\Psi_{k,l}^r + \hat\Psi_{k,l}^{r+1},\\
	a_{k,l}^p = \cos(\theta_p)\sin\left( 2\pi\frac{k-1}{M} \right) + \sin(\theta_p)\sin\left( 2\pi\frac{l-1}{M} \right).	
\end{gather*}
We refer to \cite{Remizov2013} for details.

Each of the evolution equations \eqref{eq:discr-eq} can be independently solved via standard numerical
semi-implicit schemes, recommended for this type of equation (see
\cite[Chapter~5]{Marchuk1982}).

\subsection{The reconstruction algorithm}
\label{sub:the_algorithm}

The algorithm for image inpainting via hypoelliptic diffusion is divided in three steps:
\begin{enumerate}
	\item Lift the image $f(x_k,y_l)$ to $\of(x_k,y_l,\theta_r)$.
	\item Evolve $\of(x_k,y_l,\theta_r)$ according to \eqref{eq:discr-eq} 
        after passing to the frequency grid: $\bar f \mapsto \hat{\bar f}$.
	This step was already discussed in 
        Section~\ref{ssub:two_distinct_points_of_view_leading_to_similar_computations}.
	\item Go back to the spatial grid by inverse discrete Fourier transform and project the result back 
        to the original 2-dimensional grid.
\end{enumerate}

\subsubsection{Lift}

The discrete analogue of the initial function $\of$ defined in \eqref{31}, \eqref{32} has the form:
\begin{equation}
\label{5}
\of (x_k,y_l, \theta_r) =
 \left \{
 \begin{aligned}
 f(x_k,y_l), \ \ \,  &\textrm{if} \ \  \theta_r \simeq \theta (k,l),   \\
 \phantom{0}0,\phantom{000}  \ \ \,                     &\textrm{if} \ \  \theta_r \not \simeq \theta (k,l).   \\
 \end{aligned}
 \right.
\end{equation}
Here, $\theta (k,l)$ is the discrete analogue 
of the slope angle of the level curve $\{f(x,y)=\const\}$ passing through the point $(x_k,y_l)$, that is, 
\begin{equation}
\tan \theta (k,l) = -\frac{f_x}{f_y}(x_k,y_l),  
\label{6}
\end{equation}
where $f_x$ and $f_y$ are the standard finite-difference analogues of the corresponding partial derivatives.
The notation $\theta_r \simeq \theta (k,l)$ means that $\theta_r$ is 
the nearest point to $\theta (k,l)$ among all points of the grid 
$\{ \theta_0, \ldots, \theta_{N-1} \}$ (any of  nearest points if there are two).

If $f_x(x_k,y_l)=f_y(x_k,y_l)=0$ (which corresponds to a critical point of the function $f$) we define
\begin{equation}
\label{7}
\of (x_k,y_l, \theta_r) =  \frac{f(x_k,y_l)}{N}  \quad  \text{for} \ \  \ r=0, \ldots, N-1.
\end{equation}
Generically, due to the Morse property,  $|f_x| + |f_y| \neq 0$ at almost all points $(x_k,y_l)$ 
and the function $\of$ is defined by  formulae \eqref{5} and \eqref{6}. 
Thus the information about the initial image is contained in $\theta (k,l)$ and $f(x_k,y_l)$. 

In practice, calculation of the slope angle $\theta (k,l)$ can have a large error appearing due to 
corrupted pixels, especially in the case of highly corrupted images (for instance, presented in Fig.~\ref{qqq8}). 
Therefore, it is important to know how does the distortion of this information affect the reconstruction. 
Section~\ref{sub:new_exp} contains a series of experimental results answering this question.

\subsubsection{Projection}
\label{sub:projection}

The final step of our algorithm is to convert the result of the evolution \eqref{1}, denoted by
\begin{equation*}
\oF(x,y, \theta_r)=\psi(x,y, \theta_r,T)
\end{equation*}
into a function $F(x,y)$, which represents the reconstructed image. 
Observe that, due to the well-known properties of the (hypoelliptic) heat evolution, 
and the fact that the initial function $\of$ is non-negative, the same is true for $\psi$ at any time $t>0$. 
(See, e.g., \cite{Strichartz, Strichartz-corr}.) 

A natural choice for this projection procedure is to consider the $\ell^p$-norm of the function $\oF(x,y, \theta)$ with respect to $\theta  \mod \pi$, where  $1 \le p \le \infty$.
As discussed in \cite{Remizov2013}, we have chosen the $\ell^{\infty}$ norm. That is, 
\begin{equation}
	\label{8}
	F(x_k,y_l) = \max_{\theta_r} \oF(x_k,y_l, \theta_r). \\
\end{equation}

After the projection, we obtain a non-negative function $F(x,y)$, whose maximal value, due to
the action of the diffusion, is usually small.  Therefore, it is necessary to renormalize: 
\begin{equation*}
F(x_k,y_l)  \mapsto  \frac{F(x_k,y_l)}{\max\limits_{k,l} F(x_k,y_l)}.
\end{equation*}

\begin{figure*}[t]
\begin{center}
\subfigure{\includegraphics[width=3.75cm]{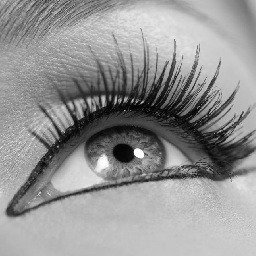}}
\subfigure{\includegraphics[width=3.75cm]{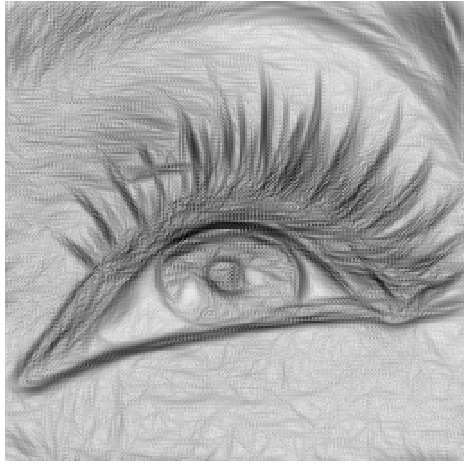}}
$\qquad$
\subfigure{\includegraphics[width=3.75cm]{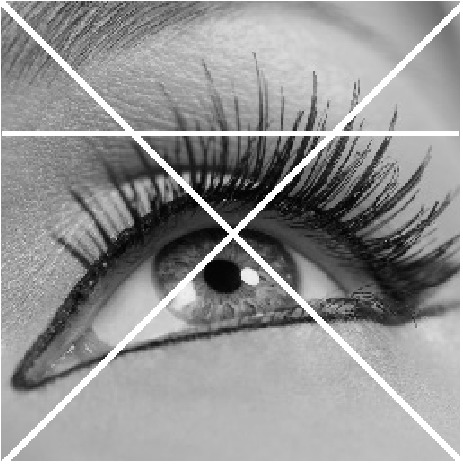}}
\subfigure{\includegraphics[width=3.75cm]{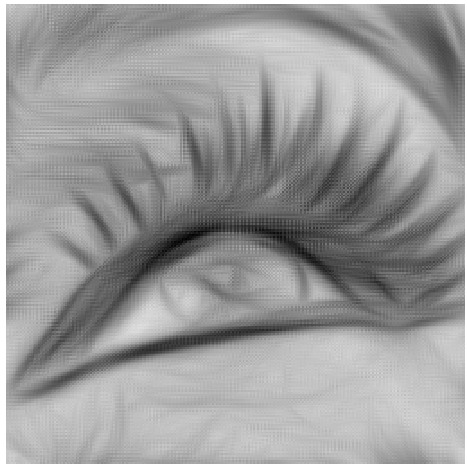}}
\caption{
The evolution of the diffusion with different final times.
The left pair presents the diffusion of non-corrupted image with small final time.  
The right pair presents the diffusion of slightly corrupted image with larger final time $T$ necessary for filling the white strips.
In the both cases, the lift is done by \eqref{5}--\eqref{7}
}
\label{qqq1}
\end{center}
\end{figure*}

\begin{figure*}
\begin{center}
\subfigure{\includegraphics[width=16.0cm]{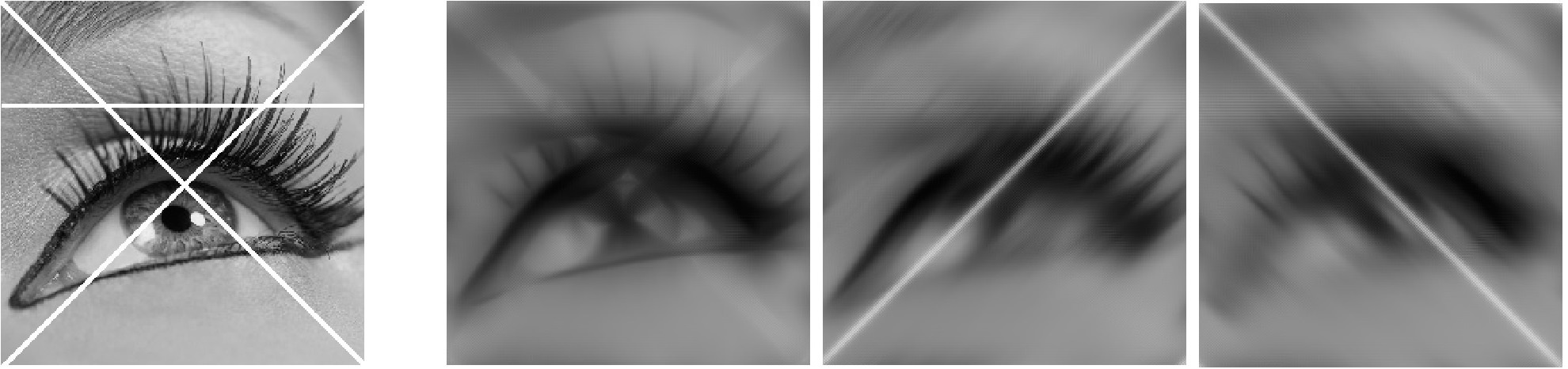}}
\caption{
From the left to right: the corrupted image and 
three processed images showing the anisotropic nature of the diffusion. 
The lift is taken to be, respectively, the trivial one given by \eqref{7} at all points of the image, 
a lift with constant slope angle $\theta(k,l) \equiv \frac{\pi}{4}$ and 
a lift with constant slope angle $\theta(k,l) \equiv \frac{3\pi}{4}$.
}
\label{qqq2}
\end{center}
\end{figure*}

\begin{figure*}
\begin{center}
\subfigure{\includegraphics[width=3.75cm]{eye-non-corr}}
\subfigure{\includegraphics[width=3.75cm]{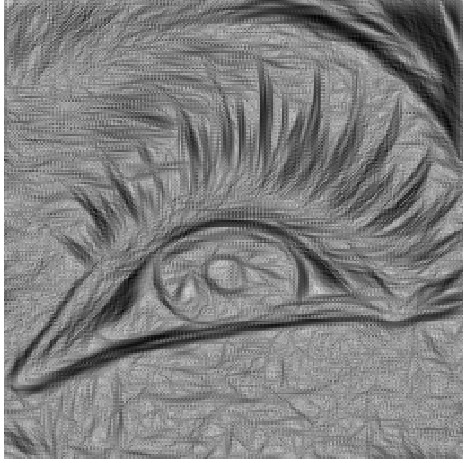}}
$\qquad$
\subfigure{\includegraphics[width=3.75cm]{eye}}
\subfigure{\includegraphics[width=3.75cm]{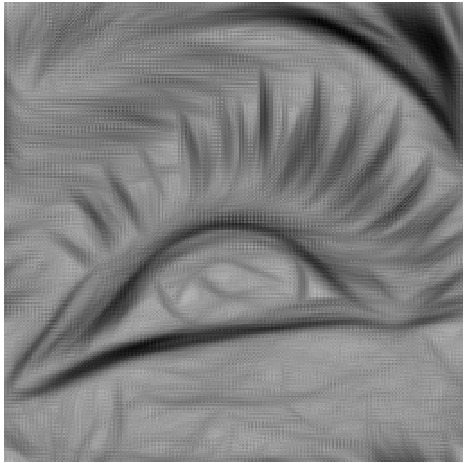}}
\caption{
Two reconstructions showing the anisotropic nature of the diffusion. 
In both cases, the lift is done by formula \eqref{5}, where the value $f(x_k,y_l)$ is replaced by the constant $\frac12$. 
The left pair presents the diffusion of non-corrupted image with small final time.  
The right pair presents the diffusion of slightly corrupted image with larger final time $T$ necessary for filling the white strips.
}
\label{qqq3}
\end{center}
\end{figure*}


\subsection{Numerical experiments}
\label{sub:new_exp}

In Fig.~\ref{qqq1} to \ref{qqq3} we present some experimental results related to the above mentioned methods. They concern only the hypoelliptic evolution, without the SR/DR procedures discussed in the next section. 
{As already mentioned, these experiments are done on images of size $256\times 256$ and $N=30$. No improvement is visible by choosing $N>30$.}  

\begin{itemize}
	\item Fig.~\ref{qqq1} presents the evolution of the diffusion in time. 
	The initial image is lifted according to \eqref{5}--\eqref{7}.
	\item Fig.~\ref{qqq2} shows the anisotropic effect of the diffusion:
	The three processed images correspond to different kinds of lift.
	The first one is obtained with the trivial lift \eqref{7}.
	The second processed image corresponds to a lift with the constant angle $\tfrac{\pi}{4}$ only, while the last one corresponds to a lift with the constant angle $\tfrac{3\pi}{4}$.  Observe how in the two latter cases, the diffusion completely fills the white lines transversal to the fixed direction and preserves the one parallel to it.
	\item Fig.~\ref{qqq3} 	illustrates the effect of the following perturbation of the lift: 
	 in \eqref{5} the slope angle  $\theta_r \simeq \theta (k,l)$  is properly calculated by  \eqref{6}, 
	 but the true value $f(x_k,y_l)$ is replaced with a non-zero constant.
\end{itemize}

In Fig.~\ref{qqq2} and \ref{qqq3} we show how modifying the lifting procedure alters the results of the diffusion, which however keeps its anisotropic character. Comparing these images with Fig.~\ref{qqq1}, one can see that the diffusion gives best results if the lift is obtained via \eqref{5}--\eqref{7}. {However, the trivial lift given only by \eqref{7} is useful when treating highly corrupted images, for which the precise evaluation of the gradient necessary to apply \eqref{5} is unachievable. Thus, in the following, we will always consider the trivial lift when the corruption is higher than $80\%$.}

\begin{figure*}
\begin{center}
\begin{minipage}{.45\textwidth}
\includegraphics[width=\textwidth]{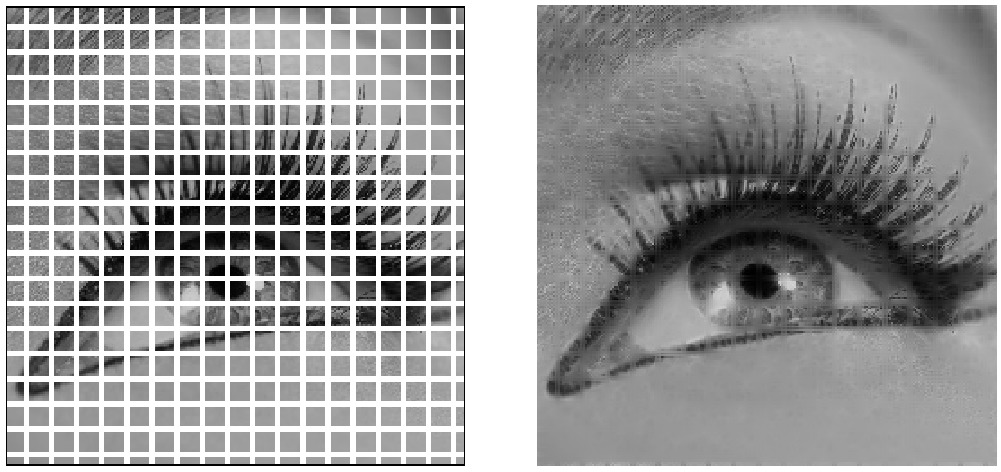}
\label{eye311}
\end{minipage}
\qquad
\begin{minipage}{.45\textwidth}
\includegraphics[width=\textwidth]{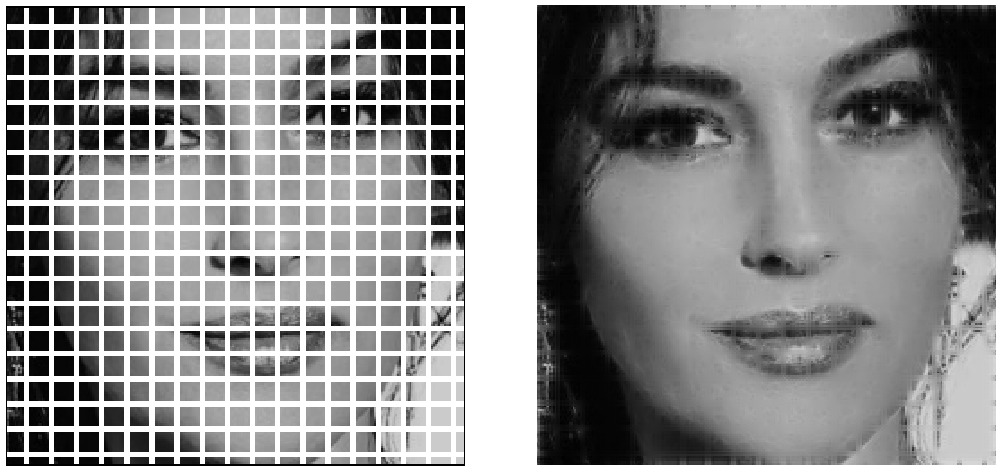}
\label{bellucci311}
\end{minipage}
\begin{minipage}[b]{.45\textwidth}
\includegraphics[width=\textwidth]{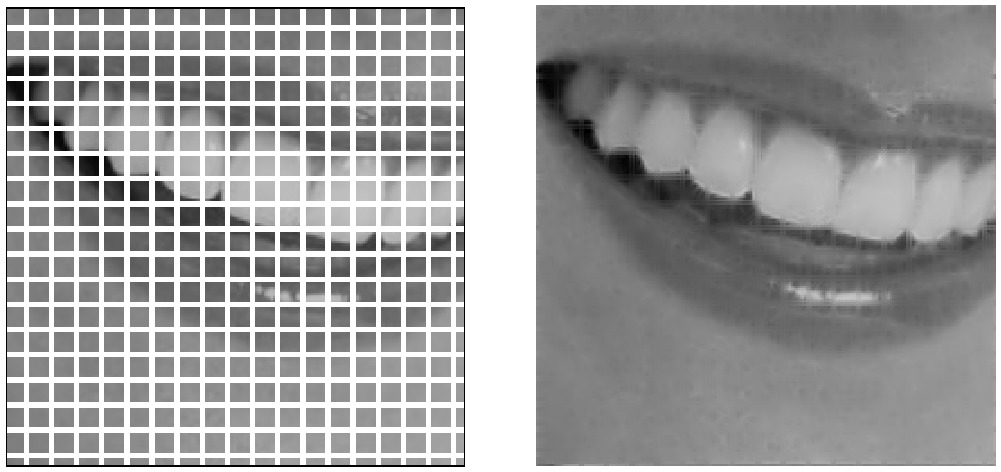}
\label{smile311}
\end{minipage}
\qquad
\begin{minipage}[b]{.45\textwidth}
\includegraphics[width=\textwidth]{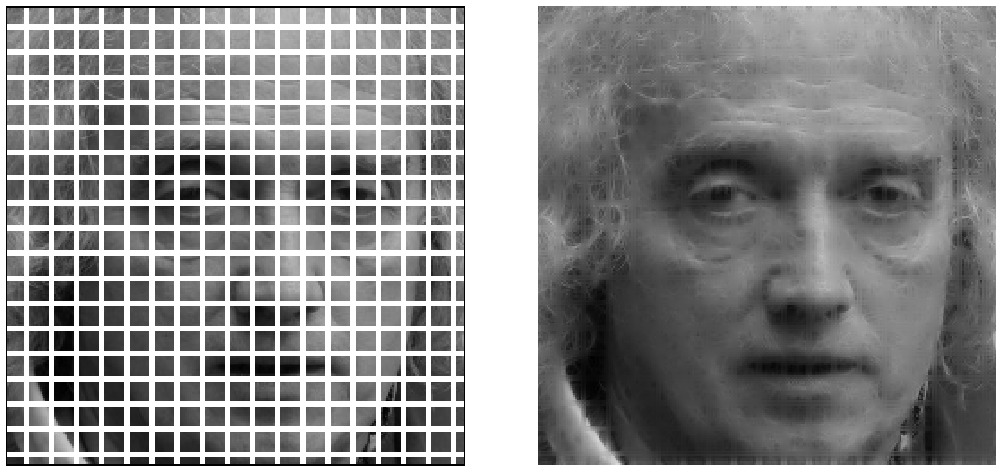}
\label{jpg311}
\end{minipage}
\caption{
Images reconstructed with the hypoelliptic equation with varying coefficients 
and the DR procedure, Section~\ref{var_coeff}.
Total corruption: 37\%, width of corrupted lines: 3 pixels.
}
\label{qqq4}
\end{center}
\end{figure*}

\subsection{Heuristic complements: SR/DR procedures}
\label{sub:heuristic_complements}

The above procedure has the drawback of applying the evolution to the whole image, and
thus also on the non-corrupted part, blurring it.  In \cite{Remizov2013}, we proposed an
heuristic complement, allowing to keep track of the initial information during the
evolution.  This method is based upon the general idea of distinguishing between the
so-called {\it good} and {\it bad} points (pixels) of the image under reconstruction.
Roughly speaking, the set $G$ of good points consists of points that are already
reconstructed enough (thus including non-corrupted points), while the set $B$ of bad
points consists of points that are still corrupted.  This procedure then amounts to
slowing the effects of the diffusion on the set $G$, without influencing~$B$.
The idea of the restoration procedure is to ``mix'' the solution $\psi(x,y,\theta,t)$ of the diffusion equation 
with the initial function $\psi(x,y,\theta,0)={\overline f}(x,y,\theta)$ at each point $(x,y) \in G$.

In \cite{Remizov2013}, we described two possible realizations of this idea 
called {\it static restoration} (SR) and {\it dynamic restoration} (DR).  The difference between the
SR and the DR procedure consists in the way the sets of good and bad points are handled:
in the SR procedure the set of good points $G$ coincides with the set of non-corrupted points
and does not change during the diffusion, while in the DR procedure $G$ 
coincides with the set of non-corrupted points only initially and bad points are allowed
to become good through the action of the diffusion.

An example of an image reconstructed with the DR procedure is given in Fig.~\ref{fig:comparison}(b). In Fig.~\ref{fig:comparison}(c) and (d), the same image is reconstructed with the more efficient methods presented in
the following sections. As it will be explained later, the method used to obtain Fig.~\ref{fig:comparison}(c) is a combination of the DR procedure with the modified hypoelliptic diffusion presented in the next section.


\section{A first improvement: Hypoelliptic diffusion with varying coefficients}\label{var_coeff}

One can try to modify equation \eqref{1} to take more into account the knowledge of the corrupted part of the image.  
A natural idea is to apply the diffusion only to corrupted regions of the image or to apply it with different final time $T$, 
larger at corrupted pixels and smaller at non-corrupted pixels. This approach requires the decomposition of the images into 
different domains with the subsequent reconciliation of the results of the diffusion. This scheme is very difficult to realize in practice. 
Therefore, we chose a different approach.  

First, remark that the diffusion given by \eqref{1} has two parameters: the coefficient $\beta$ appearing in the operator $\Del_H$ and 
the final time $T$. Obviously, the initial value problem \eqref{1} is equivalent to 
\begin{equation}
\label{100}
 \left \{
 \begin{aligned}
 &
\frac{\pa \uu}{\pa t} = \Del_H \uu, \ \ \, \Del_H   = a(X_1)^2 + b (X_2)^2,   \\
& 
\uu \I_{t=0} = \of (x,y, \theta),   \\
 \end{aligned}
 \right.
\end{equation}
where the final time is equal to $1$, and the vector fields $X_1$, $X_2$ are defined in \eqref{0}. 
Here $a,b$ are given by $a=T$ and $b = T \bet$.  

Exploiting \eqref{100}, we can control the intensity of the diffusion as a function of the position $(x,y)$, 
by considering $a,b$ as functions of $(x,y)$. 
Roughly speaking, we choose smaller values of $a,b$ at non-corrupted points and larger values at corrupted points.

The price we have to pay is the loss of the essential decoupling effect that allows to pass from
\eqref{eq:semi-discr-evolution} to the decoupled system \eqref{eq:discr-eq}.  
To overcome this point we use a well-known trick: 
at each step of integration we replace the varying coefficients equation \eqref{100} 
with a similar equation with constant coefficients. (See, e.g., \cite[Chapter~6]{ferziger}.)
Namely, let $[t_i,t_{i+1}]$ be the time interval of the integration and
consider as initial datum the function $\psi_{i}:=\psi(x,y,\theta,t_{i})$ calculated at
the previous step $[t_{i-1},t_{i}]$ (or the initial datum $\ov f$ if $t_i=0$).  

Then, we replace the differential operator $\Del_H$ on the interval $[t_i,t_{i+1}]$ with the operator
\begin{equation*}
\Del'_H := a' (X_1)^2 + b' (X_2)^2 ,
\end{equation*}
where $a',b'$ are constant coefficients chosen, for instance, as 
$a' = \max a(x,y)$, $b' = \max b(x,y)$. 
The following approximation holds
\begin{equation*}
\Del_H \psi \approx \Del'_H \psi - \Del'_H \psi_{i}  +  \Del_H \psi_{i}  = \Del'_H \psi  + d_i, 
\end{equation*}
where $d_i = \Del_H \psi_{i} - \Del'_H \psi_{i}$ can be explicitly computed.
Indeed, the approximate equality in the above formula become exact if $\psi_{i}$ is replaced with $\psi$, whence 
the approximation error is $\Del (\psi_{i}-\psi)$, where $\Del = \Del_H -\Del'_H$, is small if $t_{i+1}-t_i$ is small enough. 

Thus, on the interval $[t_i, t_{i+1}]$ we replace equation \eqref{100} with the inhomogeneous equation
\begin{equation}
\label{2a}
\frac{\pa \psi}{\pa t} = \Del'_H \psi + d_i, \ \ \ t \in [t_i, t_{i+1}], 
\end{equation}
with constant coefficients $a', b'$ and source $d_i$.
After that, the decoupling effect mentioned in
Section~\ref{ssub:two_distinct_points_of_view_leading_to_similar_computations} persists
and the semi-implicit method is still pertinent applied to each of the decoupled evolution equations, which 
differ from \eqref{eq:discr-eq} only via $d_i$. 

As already mentioned above, when choosing the varying coefficients $a, b$, the idea is to make them 
larger at bad points and their neighbors (especially the coefficient $a$, which has the
most influence to the velocity of the diffusion). Since the bad points correspond to the
set $f(x,y)=0$, the coefficients $a(x,y)$ and $b(x,y)$ can be chosen to be a continuous 
approximation of the indicator function of the set $\{f(x,y)=0\}$. 
The continuity is desirable for better stability of the numerical integration. 
For instance, we consider the following simple formula for the coefficients:
\begin{equation}
\label{9}
	\begin{split}
	a(x,y) &= a_0 + a_1 \exp \biggl(-\frac{f^2(x,y)}{\sigma}\biggr),\\
	b(x,y) &= b_0 + b_1 \exp \biggl(-\frac{f^2(x,y)}{\sigma}\biggr),
	\end{split}
\end{equation}
where $a_i, b_i, \sigma$ are positive constant parameters chosen experimentally.

\subsection{Numerical experiments}

In Fig.~\ref{qqq4}, we present a series of reconstructions obtained with the diffusion \eqref{100}
with varying coefficients coupled with the DR procedure and using the trivial lift, i.e., 
the lift defined by \eqref{7} at all points of the image.
The coefficients of the diffusion are defined by \eqref{9} with parameters 
$a_0=0.1, \ a_1=0.4$, 
$b_0=1.1, \  b_1=10$, 
\ $\sigma = 0.1$.

{
In Fig.~\ref{fig:comparison}, we present a comparison of an image reconstructed with this method, Fig.~\ref{fig:comparison}(c), with the methods presented in the previous section, Fig.~\ref{fig:comparison}(b), and with
the final algorithm, Fig.~\ref{fig:comparison}(d), presented in the next section.
}


\section{AHE algorithm}
\label{sec:synthesis}

In this section we present the main subject of this paper: the Averaging and Hypoelliptic Evolution (AHE) algorithm.
The main idea behind the AHE algorithm is to provide the anisotropic diffusion with better initial conditions.  More
precisely, it is divided in the following 4 steps:
\begin{enumerate}
	\item Preprocessing phase (Simple averaging);
	\item Main diffusion (Strong smoothing);
	\item Advanced averaging;
	\item Weak smoothing.
\end{enumerate}

Let us denote the sets of good and bad points by respectively $G$ and $B$.  
Initially (before starting the algorithm) these sets are 
\begin{equation*}
	\begin{split}
G &=\{(x_k,y_l)\mid f(x_k,y_l)>0\}, \\  
B &= \{(x_k,y_l)\mid f(x_k,y_l)=0\},
	\end{split}
\end{equation*}
Observe that $B\cup G$ covers the whole image (the whole grid) and neither $B$ nor $G$ are empty. 
For each $(x_k,y_l) \in B$ denote by $\Theta_{kl}$
its 9-points neighborhood.  Define the set $G_{kl} = G \cap \Theta_{kl}$ and let
$|G_{kl}|$ be the cardinality of $G_{kl}$. Obviously, $0 \le |G_{kl}| \le 8$.  
We  call $\partial B$ the set of {\it boundary bad points}, i.e., of those $(x_k,y_l) \in B$ satisfying the
condition $|G_{kl}|>0$.

\begin{remark}
{  
The AHE algorithm includes the hypoelliptic diffusion with the varying coefficients presented in Section~\ref{var_coeff} (steps 2 and 4). 
At the both steps, the diffusion can be performed either with the  SR/DR procedure or without it. Numerous experiments show that 
using the SR/DR procedure allows to slightly improve the quality of reconstruction if the cardinality of the set $G$ 
(the number of non-corrupted pixels) is large enough. 
However, in the case of highly corrupted images (such as those presented in Fig.~\ref{qqq7}, \ref{qqq8})
using the  SR/DR procedure gives almost no significant improvement. 
For this reason, in the following we use the diffusion without the SR/DR procedure.
}
\end{remark}

\begin{figure*}[t]
	\begin{center}
\includegraphics[width=.75\textwidth]{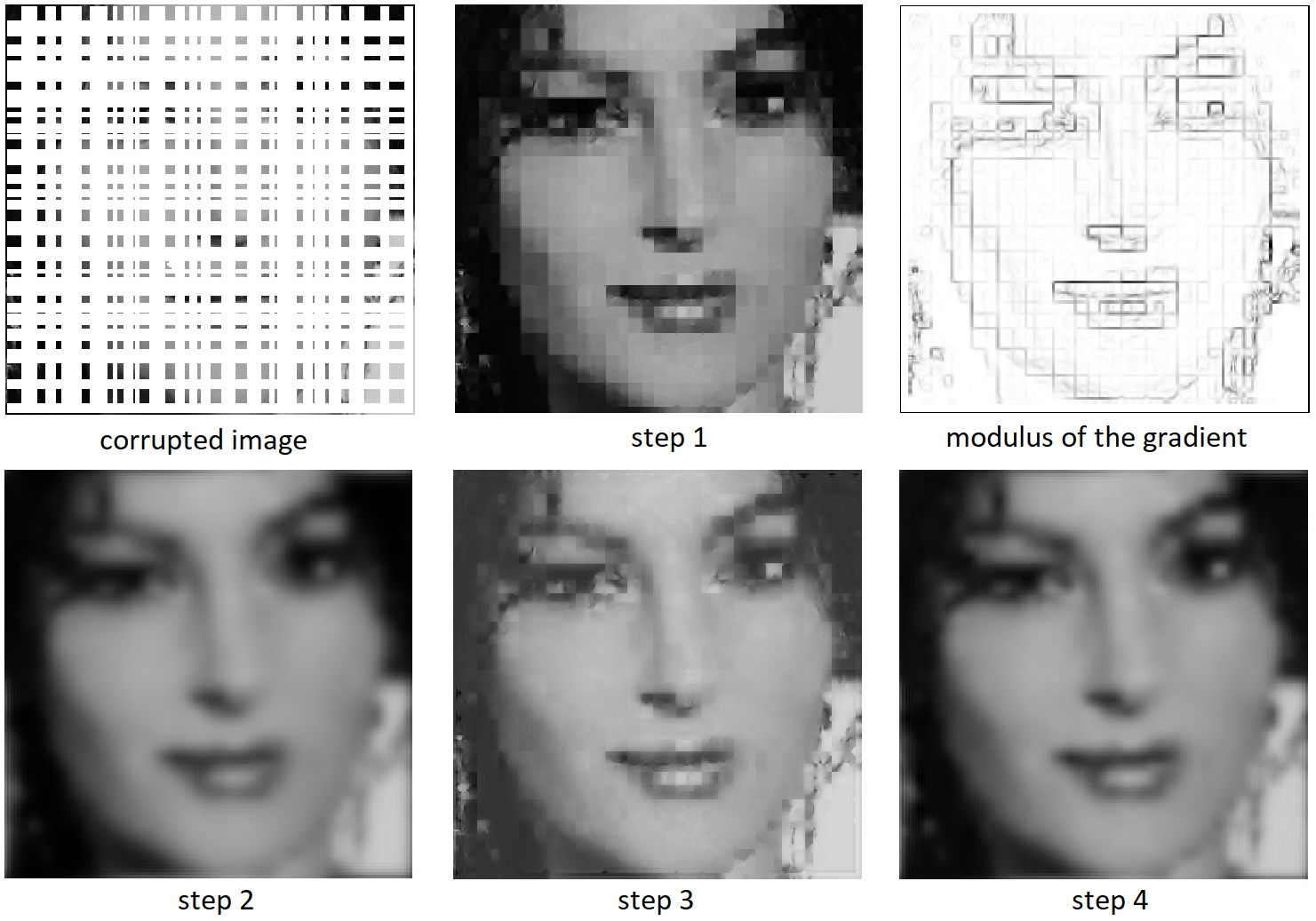}
\caption{
The result  of reconstruction  obtained after each of four steps in the AHE algorithm. 
The third image depicts the modulus of the gradient of the result of step~1, which we use to compute the varying coefficients in step~2.
}
\label{qqq5}
\end{center}
\end{figure*}

\subsection{Step 1: Preprocessing phase (Simple averaging)}

The aim of this phase is to fill in the corrupted areas of the picture with a rough
approximation of what the reconstruction should be, obtained via a discrete approximation of an isotropic  diffusion.  Namely, we iteratively redefine the value of $f$ at each boundary
bad point $(x_k,y_l)$ to be the average value of the good points in its 9-points
neighborhood $\Theta_{kl}$. Then, we remove $(x_k,y_l)$ from $B$ and add it to $G$.

More precisely, let $f^0=f$, $G^0=G$ and $B^0=B$.
Given $f^i$, $G^i$ and $B^i$ we define $f^{i+1}$, $G^{i+1}$ and $B^{i+1}$ as follows.
For any $(x_k,y_l)\in \partial B^i$ we put
\begin{equation}
f^{i+1}(x_k,y_l) =
{\frac{1}{|G^{i}_{kl}|} \sum_{(x,y) \in G^{i}_{kl}} }f^{i}(x,y),
\label{2-1}
\end{equation}
and for any $(x_k,y_l)\notin \partial B^i$ we put
\begin{equation*}
f^{i+1}(x_k,y_l)=f^i(x_k,y_l).
\end{equation*}
Observe, in particular, that this formula leaves the values of 
$f^{i+1}$ on $B^i\setminus \partial B^i$ to be zero.
Finally, we let $G^{i+1} = G^i\cup \partial B^i$ and $B^{i+1}=B^i\setminus \partial B^i$.

Since the set $\partial B^i=\emptyset$ if and only if $B^i=\emptyset$, 
after a finite number of step $s$ we obtain $B^s = \emptyset$.
We then let $g = f^s$ to be the result of this procedure.
Observe that, in particular, $g(x_k,y_l)>0$ for all $(x_k,y_l)$.

\subsection{Step 2: Main diffusion (Strong smoothing)}

The goal of this step is the elimination (or at least weakening) of the ``mosaic'' effect
resulting from the previous step.  Here, we apply diffusion~\eqref{100} with  
varying coefficients $a,b$ chosen so that the diffusion is more intensive at the points where the
``mosaic'' effect is more strong.  
To estimate the intensity of the ``mosaic'' effect, we use the absolute value of the
gradient of the function $g$.  Indeed, comparing the images presented in Fig.~\ref{qqq5}, 
one can see that most of the points with strong ``mosaic'' effect
coincide with the points where $|\nabla g(x,y)|$ is large.

Thus, we apply the hypoelliptic diffusion \eqref{100} with initial condition $\og  (x,y, \theta)$, obtained from $g(x,y)$ 
by the trivial lift \eqref{7} at all points. 
The choice of the trivial lift has an obvious advantage if we deal with highly corrupted images: 
if we were using \eqref{5}\,--\,\eqref{7}, the most important contribution would not be given by the contours of the image, but by
the boundaries of the ``mosaic'' effect. 
This would force the diffusion to follow such boundaries (see  the results presented in Fig.~\ref{qqq2}), 
thus preventing the smoothing effect.  

As already mentioned above, we control the intensities of diffusion~\eqref{100} via the varying
coefficients $a(x,y)$, $b(x,y)$, which can be defined by a formula similar to~\eqref{9} with an obvious difference: 
while the coefficients~\eqref{9} correspond to slowing down the diffusion at points with
large values of $f(x,y)$, now we need to slow down the diffusion at points with small values of $|\nabla g(x,y)|$.  
For instance, 
\begin{equation}
\label{2-2}
\begin{split}
a(x,y) = a_0 + a_1 \exp \biggl(-\frac{\phi^2(x,y)}{\sigma}\biggr),
\\
b(x,y) = b_0 + b_1 \exp \biggl(-\frac{\phi^2(x,y)}{\sigma}\biggr),  \\
\end{split}
\end{equation}
where 
\begin{equation*}
\phi(x,y) = 1 - \frac{|\nabla g(x,y)|}{\max\limits_{} |\nabla g(x,y)|}.
\end{equation*}
Here, $a_i, b_i, \sigma$ are constant parameters experimentally chosen. 
In all restorations via the AHE algorithm presented in this paper, we used the following values of the parameters:
$a_0=0.05$, $a_1=0.2$, $b_0=0.55$, $b_1=5$,  $\sigma = 0.4$.
From the practical point of view, the gradient $\nabla g(x,y)$ is replaced 
by  its finite-difference approximation.

\begin{figure*}
	\begin{center}
	  \begin{minipage}[b]{.45\textwidth}
	    \includegraphics[width=\textwidth]{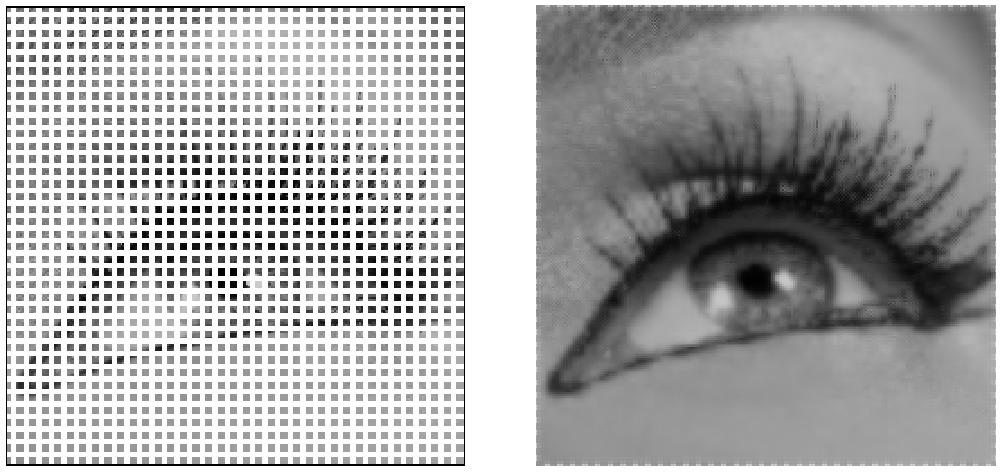}
	    \label{eye34__}
	  \end{minipage}
  	  \qquad
  	  \begin{minipage}[b]{.45\textwidth}
  	    \includegraphics[width=\textwidth]{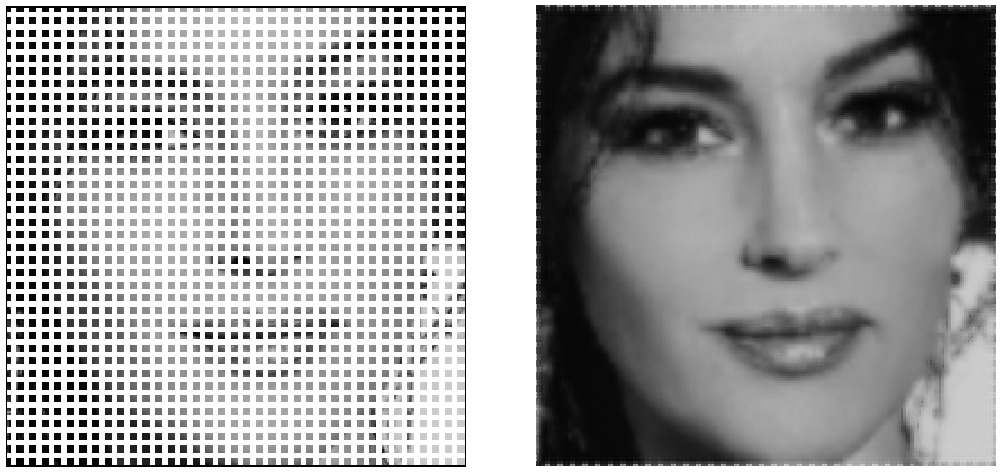}
  	    \label{bellucci34__}
  	  \end{minipage}
  	  \begin{minipage}[b]{.45\textwidth}
  	    \includegraphics[width=\textwidth]{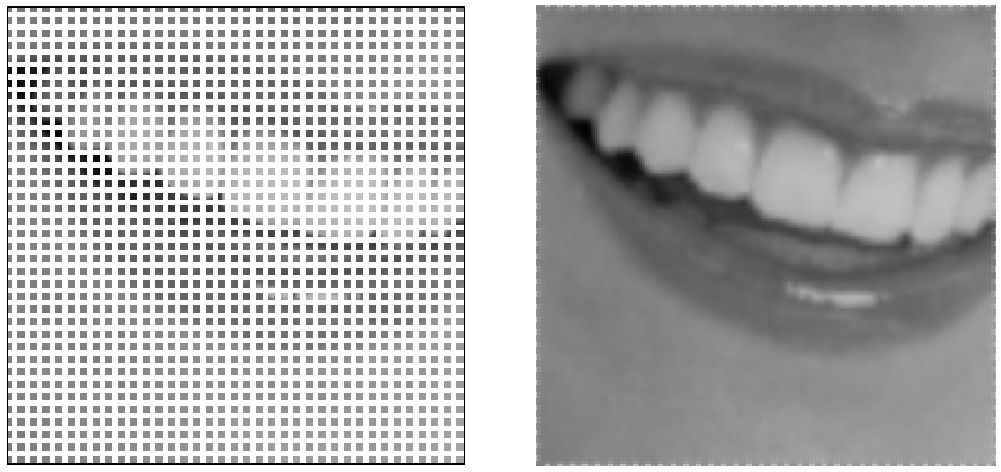}
  	    \label{smile34__}
  	  \end{minipage}
  	  \qquad
  	  \begin{minipage}[b]{.45\textwidth}
  	    \includegraphics[width=\textwidth]{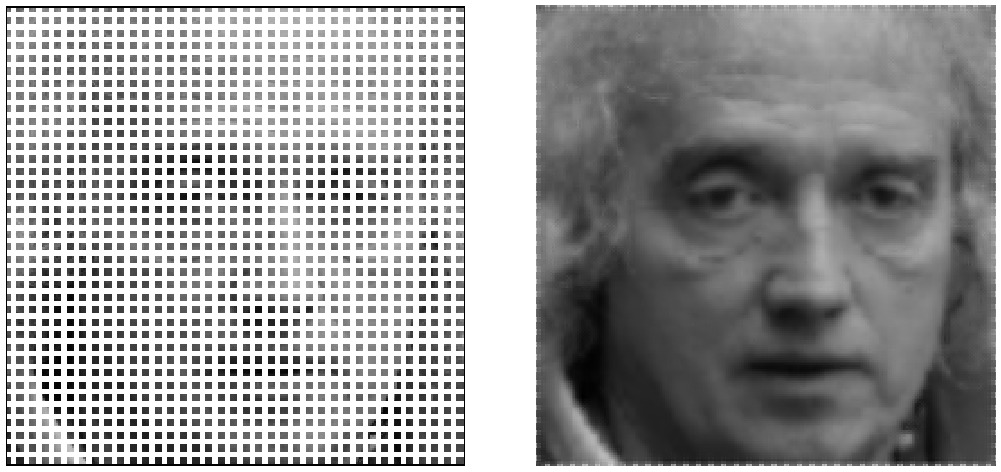}
  	    \label{jpg34__}
  	  \end{minipage}
  	  \caption{Images reconstructed with the AHE algorithm, Section~\ref{sec:synthesis}. 
          Total corruption: 67\%, width of corrupted lines: 3 pixels.}
  	  \label{qqq6}
	\end{center}
\end{figure*}

\subsection{Step 3: Synthesis (Advanced averaging)}

As can be seen in Fig.~\ref{qqq5},  after the second step of the AHE algorithm we 
remove the  ``mosaic'' effect.  However, the diffusion introduces a
blurring effect, that cannot be removed by decreasing the coefficients $a,b$,
{since these have to be sufficiently large in order to remove the ``mosaic'' effect.}
To pass between this Scylla and
Charybdis, we then make a synthesis of the images obtained after the first and the second
steps.

As before, let $f(x,y)$ be the function of the initial corrupted image, and $B,G$ be the
corresponding sets of good and bad points.  Recall that we denoted by $g(x,y)$  the
function obtained after the first step and let $h(x,y)$ denote the function obtained after
the second step.

The structure of step~3 is similar to the one of step~1.  Indeed, we will apply an
iterative procedure aimed to reconstruct the bad points of $f$ using information from the
good points and the function $h$.  The only difference between steps 1 and 3 is that when
$(x_k,y_l)\in \partial B^i$, we define $f^{i+1}(x_k,y_l)$ as
\begin{equation}
f^{i+1}(x_k,y_l) = \arg \min_{X\in[0,1]}
\sum_{(x,y) \in G^i_{kl}}  \biggl| \frac{X}{f^i(x,y)} - \frac{h(x_k,y_l)}{h(x,y)} \biggr|^2.
\label{2-3}
\end{equation}
This expression realizes a compromise between the averaging and the diffusion.  The above
formula is well defined since $f(x,y)>0$ for all $(x,y) \in G^i_{kl}$ and the
smoothed function $h(x,y)$ is always strictly positive.  Moreover, the expression in
the right-hand side of \eqref{2-3} is a continuous convex function of $X$, and thus the
minimum exists.  A straightforward computation allows then to compute explicitly
\eqref{2-3} as
\begin{equation*}
f^{i+1}(x_k,y_l)  = h(x_k,y_l) \, \frac{\sum\limits_{(x,y) \in G^i_{kl}}
f^i(x,y)^{-1} h(x,y)^{-1}}{\sum\limits_{(x,y) \in
G^i_{kl}} f^i(x,y)^{-2}}.
\end{equation*}

The results of this reconstruction are presented in Fig.~\ref{qqq5}.
{As desired, we obtain a somewhat intermediate result, between step~1 and step~2.}

\subsection{Step 4: Weak smoothing}

As can be seen from Fig.~\ref{qqq5}, step~3 also reintroduces ``mosaic'' effect, but less than step~1.
Therefore, we essentially need to repeat step~2. 
The only difference is that the parameters $a_i, b_i$ in \eqref{2-2} should be chosen smaller than those in step~2. 

\begin{figure*}
\begin{center}
\begin{minipage}{.45\textwidth}
\includegraphics[width=\textwidth]{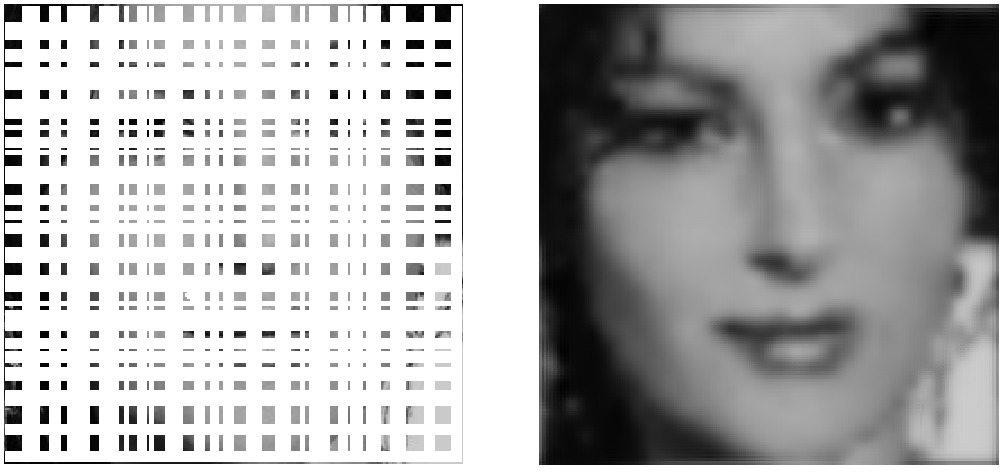}
\label{bellucci000__}
\end{minipage}
\qquad
\begin{minipage}{.45\textwidth}
\includegraphics[width=\textwidth]{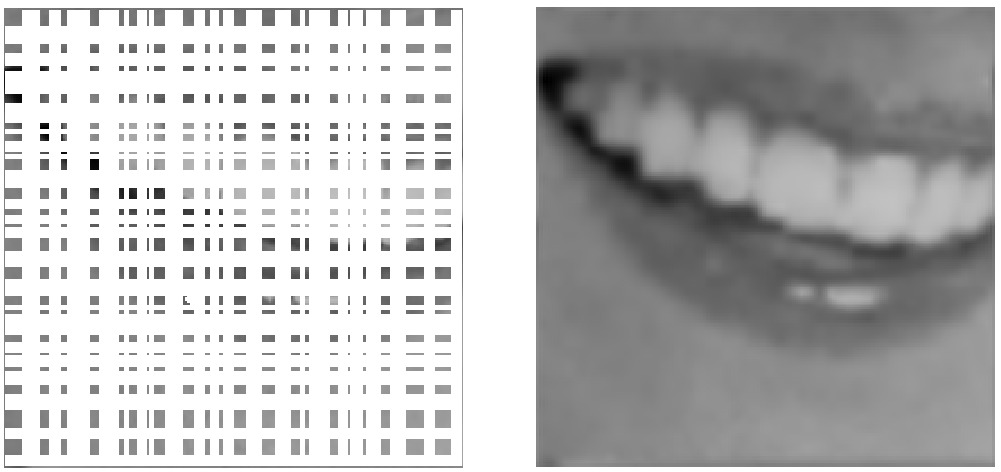}
\label{smile000__}
\end{minipage}
\begin{minipage}{.45\textwidth}
\includegraphics[width=\textwidth]{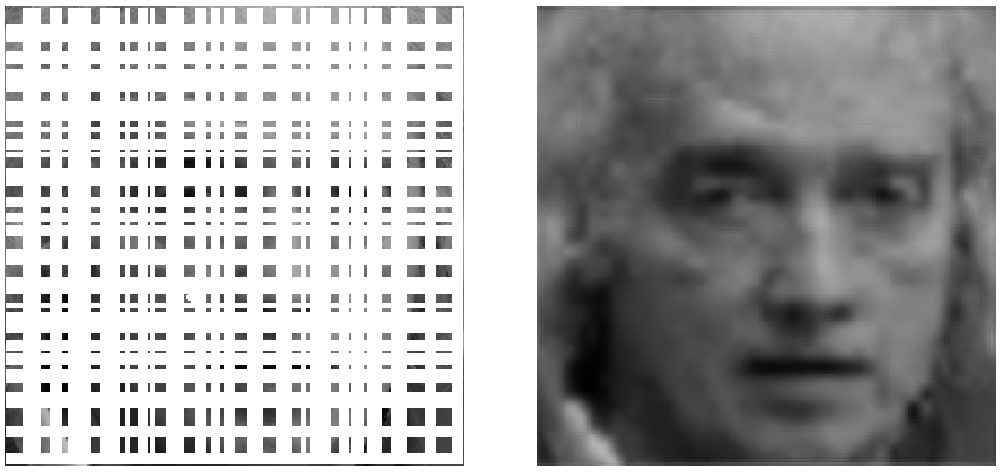}
\label{jpg000__}
\end{minipage}
\qquad
\begin{minipage}{.45\textwidth}
\includegraphics[width=\textwidth]{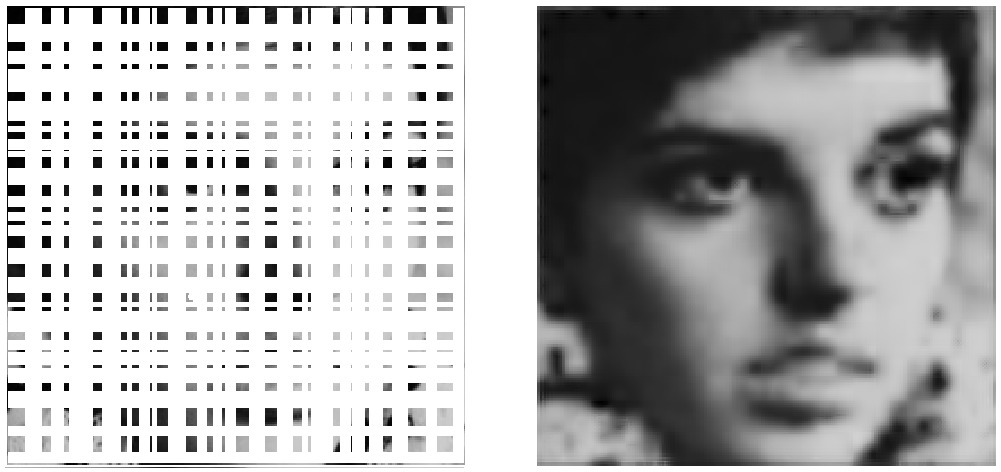}
\label{bust000__}
\end{minipage}
\caption{
Images reconstructed with the AHE algorithm, Section~\ref{sec:synthesis}.
Total corruption: 85\%.}
\label{qqq7}
\end{center}
\end{figure*}

\begin{figure*}
\begin{center}
\begin{minipage}[b]{.45\textwidth}
\includegraphics[width=\textwidth]{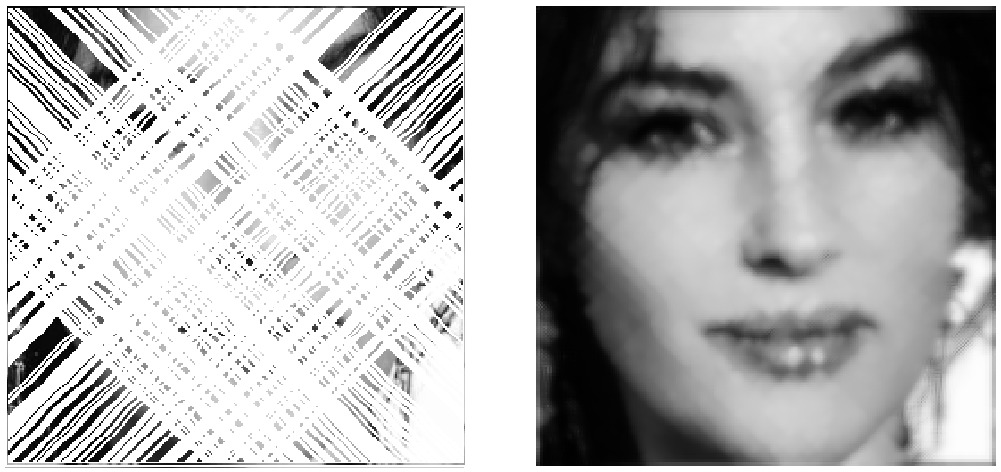}
\label{bellucci-diagonal}
\end{minipage}
\qquad
\begin{minipage}[b]{.45\textwidth}
\includegraphics[width=\textwidth]{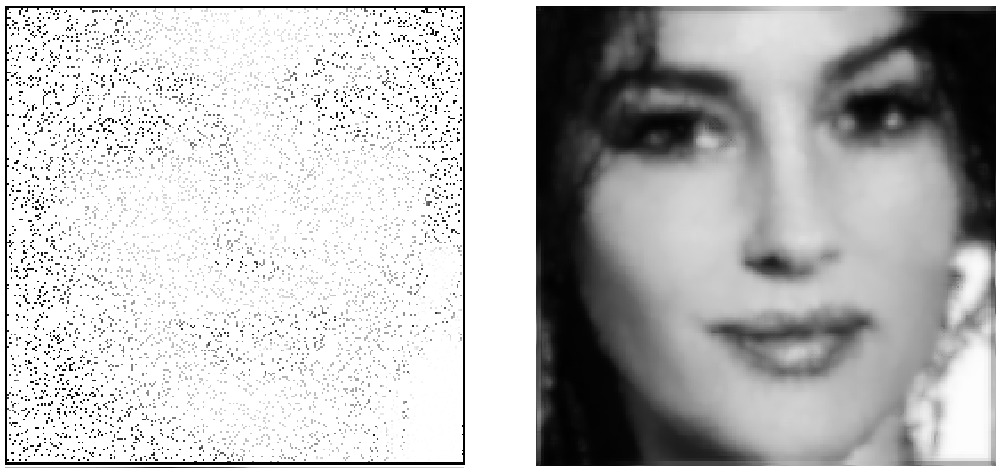}
\label{bellucci-random97}
\end{minipage}
\begin{minipage}[b]{.45\textwidth}
\includegraphics[width=\textwidth]{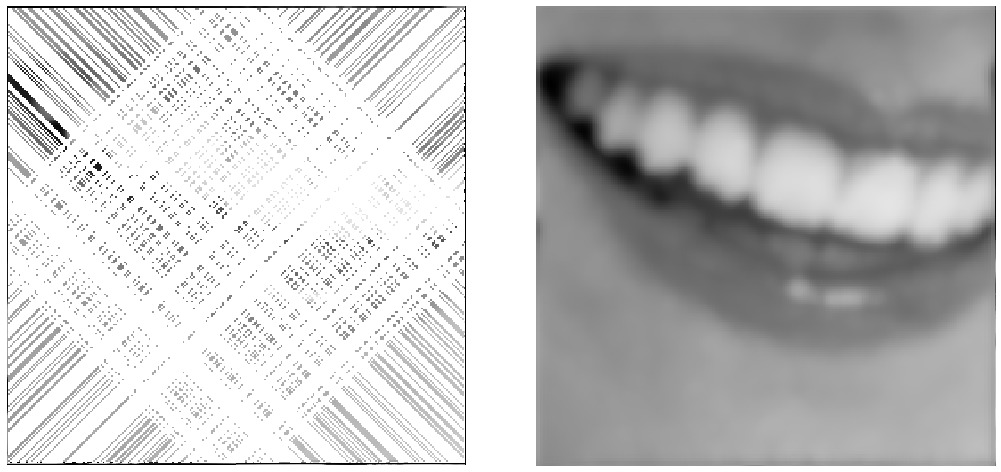}
\label{smile-diagonal}
\end{minipage}
\qquad
\begin{minipage}[b]{.45\textwidth}
\includegraphics[width=\textwidth]{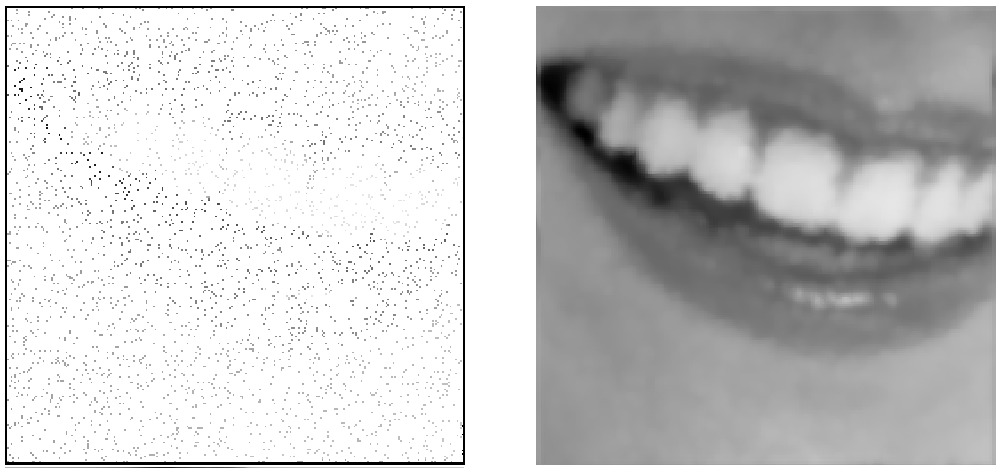}
\label{smile-random97}
\end{minipage}
\begin{minipage}[b]{.45\textwidth}
\includegraphics[width=\textwidth]{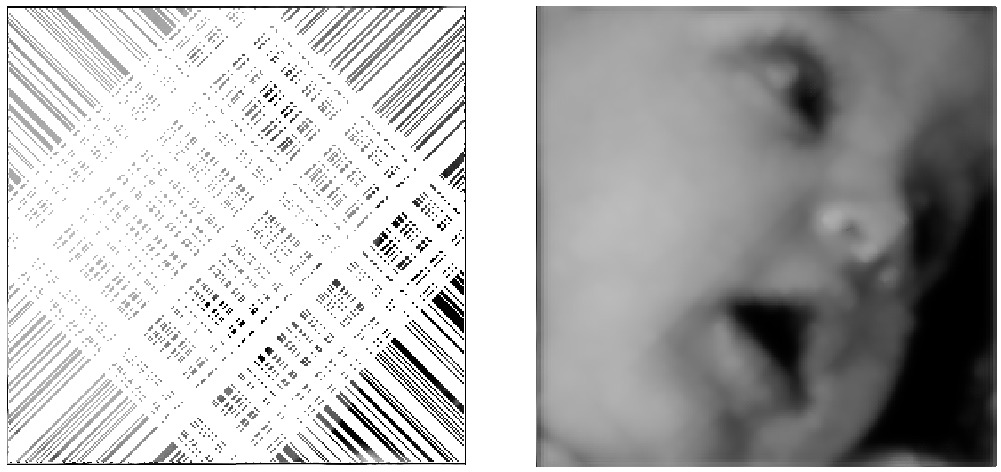}
\label{elettra-diagonal}
\end{minipage}
\qquad
\begin{minipage}[b]{.45\textwidth}
\includegraphics[width=\textwidth]{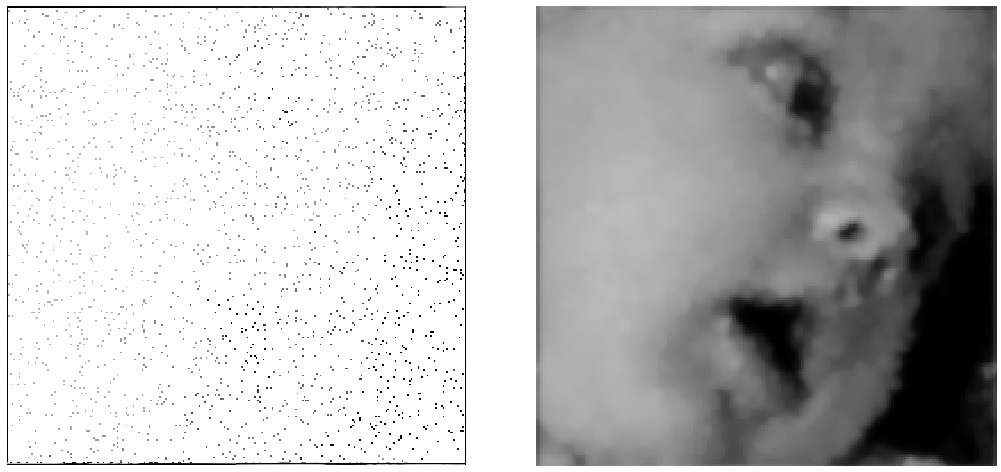}
\label{elettra-random97}
\end{minipage}
\caption{
Images reconstructed with the AHE algorithm, Section~\ref{sec:synthesis}.
Two types of corruption are presented here. 
On the left: diagonal lines, total corruption is about 80\%. 
On the right: uniform random distribution of corrupted pixels, total corruption is 90\% (Bellucci), 95\% (smile), 97\% (child).
}
\label{qqq8}
\end{center}
\end{figure*}

In all reconstructed images presented in Fig.~\ref{qqq6}\,--\,\ref{qqq8}  we use the trivial lift \eqref{7} and
hypoelliptic diffusion~\eqref{100} with varying coefficients $a,b$ defined by~\eqref{2-2}.
For the results presented in Fig.~\ref{qqq7}, \ref{qqq8}, we used the following parameters: 
$a_0=0.015$, $a_1=0.1$, $b_0=0.15$, $b_1=1.5$,  $\sigma =0.3$.

\subsection{Numerical cost of the algorithm}
\label{sec:complexity}

Computational cost of the AHE algorithm is moderate. Let us consider an
input image of size $M\times M$ pixels and $N$ possible directions. The
most computationally expensive part is the hypoelliptic diffusion with varying coefficients, 
which appears in the AHE algorithm twice (steps 2, 4). 

At each time step, the hypoelliptic diffusion is represented by a 
system of $M^2$  linear inhomogeneous evolution equations. 
Each of them is solved using the Crank-Nicolson scheme (see, e.g.,
\cite{Marchuk1982}), which requires to solve a system of linear
algebraic equations with a $N\times N$ periodic tridiagonal matrix.
This can be done in $O(N)$ operations via a variation of Thomas algorithm.
Thus, taking into account the two-dimensional Fast Fourier Transforms (FFTs)
necessary to decouple the system, which require $O(M^2\log M)$ operations each,
the total computational cost per time step is 
\begin{equation*}
O(NM^2+NM^2\log M)=O(NM^2\log M). 
\end{equation*}

The run-time of the sequential implementation of the
AHE algorithm used to perform the reconstructions presented in this paper
is of about two minutes. The code has been run on an Intel i7-4600M CPU,
with parameters $M=256$ and $N=32$. 
We remark that the systems of $M^2$  linear inhomogeneous evolution equations are completely decoupled, as are the two-dimensional FFTs. 
This can be exploited to develop a parallel implementation, allowing {for} a significant reduction of the run-time.


\section{Conclusion}

\begin{figure*}[t]
\begin{center}
  \includegraphics[width=\textwidth]{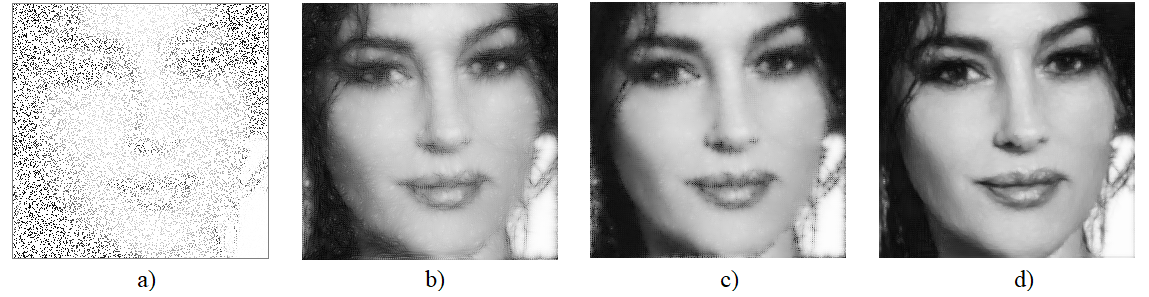}
  \caption{A comparison of reconstructions of an image with $80\%$ of pixels missing. 
  \emph{a)} the original corrupted image. 
  \emph{b)} Reconstruction with the DR method and the hypoelliptic diffusion presented in Section~\ref{sec:hypo}. See also \cite{Remizov2013}.
  \emph{c)} Reconstruction with the DR procedure and the varying coefficient hypoelliptic diffusion presented in Section~\ref{var_coeff}. 
  \emph{d)} reconstruction with the AHE algorithm.}
  \label{fig:comparison}
\end{center}
\end{figure*}

\begin{figure*}[t]
\begin{center}
  \includegraphics[width=\textwidth]{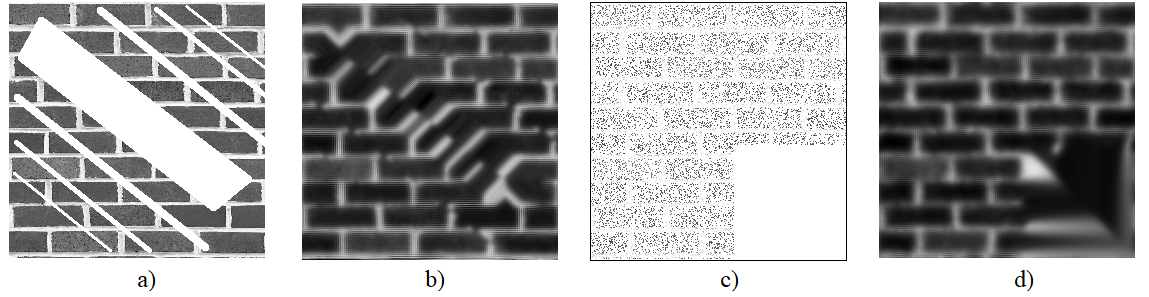}
  \caption{
  \emph{a)} Image containing small and large corrupted regions. 
  \emph{b)} Reconstruction by the AHE algorithm.
  \emph{c)} Image containing uniformly random corruption and a single large corrupted region. 
  \emph{d)} Reconstruction by the AHE algorithm.
  }
  \label{qqq10}
\end{center}
\end{figure*}

The AHE algorithm presented in the paper, provides an efficient method of reconstruction for greyscale images, including highly corrupted ones. {A comparison with the other methods presented in this paper is pictured in Fig.~\ref{fig:comparison}.}
We stress that this inpainting technique can be applied independently of the structure and the geometry of the corruption, although it requires the precise knowledge of its location. The quality of reconstruction strongly depends on the accuracy of this information. 

Notice that the effectiveness of the algorithm depends also on the distribution of the corrupted pixels. Fig.~\ref{qqq6}--\ref{qqq8} show that if the corruption is ``well distributed'' one can achieve good reconstructions even in presence of 97\% pixels missing. However, if the image contains large corrupted regions, then the reconstructions are no longer satisfactory. To this effect, see Fig.~\ref{qqq10}.

It seems obvious that the AHE algorithm is open to further development. 
For instance, the first step (simple averaging) can be replaced with a more advanced method. 
Also, the detection of the regions presenting a ``mosaic'' effect is
currently done in a very naive way, via \eqref{2-2}. Moreover, the
coefficients $a_i,b_i$ and $\sigma$ appearing in that equation have been
determined experimentally and their choice can
clearly be optimized. This step could be done, for example, via image
recognition methods based on the semi-discrete group of  
rototranslations \cite{PrandiBohi, PBG2015}.


\appendix

\section{Sub-Riemannian geometry}
\label{sec:sR}

In this Appendix we recall some standard definitions of sub-Riemannian geometry and hypoelliptic operators.  Classical texts are \cite{montgomery,nostrolibro,bellaiche,gromov}.
\begin{definition}
A $(n,m)$-sub-Riemannian manifold is given by a triple $(M,\distr,{\mathbf g})$, 
where
\begin{itemize}
\item $M$ is a connected smooth manifold of dimension $n$;
\item $\distr$ is a smooth distribution of constant rank $m< n$ satisfying the {\em H\"ormander condition}. That is, $\distr$ is a smooth map that associates to $q\in M$  an $m$-dimensional subspace $\distr(q)$ of $T_qM$, such that $\forall~q\in M$ we have
\begin{equation}
T_qM={\spn}\{[X_1,[\ldots[X_{k-1},X_k]]](q)~|~X_i\in\mathrm{Vec}_H(M)\}.
\end{equation}
Here, $\mathrm{Vec}_H(M)$ denotes the set of {\em horizontal smooth vector fields} on $M$, i.e. $$\mathrm{Vec}_H(M)=\{X\in\mathrm{Vec}(M)\ |\ X(q)\in\distr(q)~\ \forall~q\in M\}.$$
\item ${\mathbf g}_q$ is a Riemannian metric on $\distr(q)$, smooth 
as function of $q$.
\end{itemize}
\end{definition}

A Lipschitz continuous curve $q(\cdot):[0,T]\to M$ is said to be {\it horizontal} if 
$\dot q(t)\in\distr(q(t))$ for almost every $t\in[0,T]$. Given an horizontal curve $q(\cdot):[0,T]\to M$, the {\it length of $q(\cdot)$} is
\begin{equation}
\ell(q(\cdot))=\int_0^T \sqrt{ {\mathbf g}_{q(t)} (\dot q(t),\dot q(t))}~dt.
\label{e-lunghezza}
\end{equation}
The {\it distance} induced by the sub-Riemannian structure on $M$ is the 
function
\begin{equation}
d(q_0,q_1)=\inf \{\ell(q(\cdot))\mid q(0)=q_0,q(T)=q_1,
q(\cdot)\ \mathrm{horizontal}\}.
\end{equation}

The connectedness assumption for M and the H\"ormander condition guarantee the finiteness and the continuity of $d(\cdot,\cdot)$ with respect to the topology of $M$ (Chow's Theorem, see for instance \cite{nostrolibro}). The function $d(\cdot,\cdot)$ is called the {\it Carnot-Carath\'eodory distance} and gives to $M$ the structure of metric space.

Locally, the pair $(\distr,{\mathbf g})$ can be specified by assigning a set of $m$ smooth vector fields spanning $\distr$, that are moreover orthonormal for ${\mathbf g}$, i.e.  
\begin{equation}
\distr(q)={\spn}\{X_1(q),\dots,X_m(q)\},~~~{\mathbf g}_q(X_i(q),X_j(q))=\delta_{ij}.\label{trivializable}
\end{equation}
Such a  set $\{X_1,\ldots,X_m\}$ is called a {\it local orthonormal frame} for the sub-Riemannian structure. When  $(\distr,{\mathbf g})$ can be defined by $m$  globally defined vector fields  as in \eqref{trivializable} we say that the sub-Riemannian manifold is {\it trivializable}.

Given a trivializable $(n,m)$-sub-Riemannian manifold, the problem of finding a curve realizing the distance between two fixed points  $q_0,q_1\in M$ is
naturally formulated as the following optimal control problem 
\begin{equation}
\label{eq-op}
 \left \{
 \begin{aligned}
\phantom{o} &\dot q(t)=\sum_{i=1}^m u_i(t) X_i(q(t)), \ \ u_i(\cdot) \in L^\infty([0,T],\bR), \\ 
\phantom{o} &\int\limits_0^T \sqrt{\sum_{i=1}^m u_i^2(t)}~dt \, \to \, \min,\\  
\phantom{o} &q(0)=q_0, \ \ \ q(T)=q_1.\\
 \end{aligned}
 \right.
\end{equation}

\subsection{Diffusion in a sub-Riemannian manifold}
\label{sec:diffusion}

Given a sub-Riemannian manifold $(M,\distr,{\mathbf g})$ and a smooth volume $\omega$ on $M$, the sub-Riemannian heat equation is the diffusion equation: 
\begin{equation}
\label{canicola}
\partial_t \psi=\Delta_H \psi,
\end{equation}
where $\Delta_H$ is the \emph{sub-Riemannian (or horizontal) Laplacian}, defined by
\begin{equation}
  \Delta_H \phi=\diver_\omega  \grad_H \phi, \qquad \phi\in C^2(M).
\end{equation}
Here, $\diver_\omega$ is the divergence with respect to the volume $\omega$ and $\grad_H\phi$ is the {\em horizontal gradient} of $\phi$. That is, it is the unique vector field satisfying, for every $q\in M$, 
\begin{equation}
  {\g_q}(\grad_H \phi(q),v)= d_q\phi(v)\mbox{ for every }v\in\distr(q).
\end{equation}
If $\{X_1,\ldots,X_m\}$ is a local orthonormal frame, it follows that $\grad_H \phi = \sum_{i=1}^m(X_i\phi)X_i$, and thus that
\begin{equation}
  \Delta_H\phi = \sum_{i=1}^m \left( X_i^2\phi + (\diver_\omega X_i)X_i\phi \right).
\end{equation}

Thanks to the H\"ormander condition assumed  in the definition of the sub-Riemannian manifold,  the celebrated H\"ormander Theorem \cite{Hormander}, implies the following.
\begin{theorem}
The operators $\Delta_H$ (operating on functions $\phi:M\to\bR$)  and $\Delta_H-\partial_t$ (operating on functions $\psi:M\times\bR \to\bR$) are hypoelliptic.
\end{theorem}
We recall that a second order differential operator $L$ is said to be hypoelliptic if for every distribution $\phi$ defined on an open set $\Omega$ of a manifold $N$, the condition $L\phi\in{C}^\infty(\Omega)$ implies that $\phi\in{C}^\infty(\Omega)$. In particular, the hypoellipticity of $\Delta_H-\partial_t$ implies that any solution to the heat equation \eqref{canicola} on $ M\times ]t_0,t_1[$ is  smooth. 

\begin{remark}
The sub-Riemannian structure studied in this paper is the one on $PT\bR^2$ for which the distribution is given by the vector fields
\begin{equation}
X_1(q) = \cos \theta \frac{\pa}{\pa x} + \sin \theta \frac{\pa}{\pa y},
\  \  \
X_2(q) = \frac{\pa}{\pa \theta}.
\end{equation}
The metric $\g$ is then chosen such that $\{X_1,X_2\}$ are orthogonal, and ${\g}(X_1,X_1)=1$, ${\g}(X_2,X_2)=1/\beta$, for some given $\beta>0$. By taking as volume on $PT\bR^2$ the Lebesgue measure, i.e., $\omega= dx\,dy\,d\theta$, since $X_1$ and $X_2$ are divergence free, one immediately gets
\begin{equation*}
\Delta_H= (X_1)^2 + \beta  (X_2)^2.
\end{equation*}
\end{remark}

\section*{Acknowledgments}
We deeply thank G. Facciolo,  S. Masnou, and G.P.~Pa\-na\-sen\-ko  for their help.

This work was supported by the ERC POC project ARTIV1 contract number 727283, by the ANR
project ``SRGI'' ANR-15-CE40-0018, by a public grant as part of the Investissement
d'avenir project, reference ANR-11-LABX-0056-LMH, LabEx LMH (in a joint call with
Programme Gaspard Monge en Optimisation et Recherche Op\'erationnelle), by the iCODE
institute, research project of the Idex Paris-Saclay, 
by the POCI-01-0145-FEDER-006933/SYSTEC project financed by ERDF and FCT through COMPETE2020.

\bibliography{paper}
\bibliographystyle{plain}

\end{document}